\pdfoutput=1

\documentclass[11pt]{article}

\usepackage[final]{ACL2023}

\usepackage{times}
\usepackage{latexsym}

\usepackage[T1]{fontenc}

\usepackage[utf8]{inputenc}

\usepackage{microtype}

\usepackage{booktabs}       
\usepackage{amsfonts}       
\usepackage{nicefrac}       
\usepackage{xcolor}         
\usepackage{graphicx}
\usepackage{multirow}
\usepackage{float}
\usepackage{tabularx}

\newcommand{\red}{\textcolor{red}}
\newcommand{\green}{\textcolor{green}}
\newcommand{\brown}{\textcolor{brown}}

\title{Counterfactual reasoning: Testing language models' understanding of hypothetical scenarios}

\author{%
  Jiaxuan Li\\
  University of California Irvine\\
  Irvine, CA 92617 \\
  \texttt{jiaxuan.li@uci.edu} \\
  \And
  Lang Yu \\
  Meta \\
  Seattle, WA 98109 \\
  \texttt{langyu@fb.com} \\
  \And
  Allyson Ettinger \\
  University of Chicago \\
  Chicago, IL 60637 \\
  \texttt{aettinger@uchicago.edu} 
}

\begin{document}

\maketitle

\begin{abstract}
Current pre-trained language models have enabled remarkable improvements in downstream tasks, but it remains difficult to distinguish effects of statistical correlation from more systematic logical reasoning grounded on the understanding of real world. We tease these factors apart by leveraging \emph{counterfactual conditionals}, which force language models to predict unusual consequences based on hypothetical propositions. We introduce a set of tests from psycholinguistic experiments, as well as larger-scale controlled datasets, to probe counterfactual predictions from five pre-trained language models. We find that models are consistently able to override real-world knowledge in counterfactual scenarios, and that this effect is more robust in case of stronger baseline world knowledge---however, we also find that for most models this effect appears largely to be driven by simple lexical cues. When we mitigate effects of both world knowledge and lexical cues to test knowledge of linguistic nuances of counterfactuals, we find that only GPT-3 shows sensitivity to these nuances, though this sensitivity is also non-trivially impacted by lexical associative factors.\footnote{Data and code available at \url{https://github.com/goldengua/Counterfactual_Inference_LM}.}
\end{abstract}

\section{Introduction}
Reasoning plays a central role in human communication~\citep{frank2012predicting}. While language models have demonstrated remarkable capacity on downstream tasks~\citep{devlin-etal-2019-bert,radford2019language,liu2019roberta}, it remains unclear to what extent predictions generated by language models are consequences of correlation with linguistic heuristics in the context, versus robust reasoning about causal relations grounded on understanding of world knowledge. 

In this paper we leverage \emph{counterfactual conditionals} to investigate the capacity of pre-trained LMs (PLMs) to distinguish hypothetical scenarios from reality, and to examine how this interacts with models' use of existing real world knowledge as well as shallower associative cues. Counterfactuals consist of a premise which is false in the real world but true in the hypothetical world (e.g., ``If cats were vegetarians''), and an imaginary consequence of this premise (``cats would love cabbages''). Testing language models with counterfactuals allows us to use language to manipulate what is true and what is hypothetical, and to test models' ability to separate and use this information for predictions. Previous work has established the use of counterfactual scenarios to probe inference ability~\citep{qin2019counterfactual,zellers2019hellaswag,mostafazadeh2016corpus,meng2022locating,rajani2019explain,saparov2022language,frohberg2021crass,elazar-etal-2021-amnesic,rudinger2020thinking}, but the datasets lack systematic control of lexical cues and world knowledge, which makes it likely that the performance could be attributable to spurious cues in the datasets~\citep{niven2019probing}.


For our tests we draw on and adapt inputs from existing psycholinguistic experiments. We begin by testing models' ability to override existing world knowledge when the context indicates that the correct completion involves a hypothetical world (e.g., ``if cats were vegetarian, cats would love \emph{cabbages/fish}''). We test five popular PLMs, and find that models can increase their preference for counterfactual completions given counterfactual context---however, most models rely strongly on simple lexical cues. Next we control the effect of real world knowledge and lexical triggers, to test models' understanding of what counterfactual language implies about the world state. We find that most models fail to understand real-world implications of counterfactuals and largely rely on lexical triggers---with the exception of GPT-3, which shows greater sophistication, but continues to show non-trivial susceptibility to interference from lexical-associative cues. We discuss the implications and possible interpretations of these findings with respect to linguistic competence and predictive strategies of these models.


\section{Exp1: overriding world knowledge}\label{overriding-world-knowledge}

Our first experiment investigates whether LMs are able to take a counterfactual scenario and predict a counterfactual-consistent completion that contradicts general world knowledge.

\paragraph{Items} 

We draw directly on counterfactual stimuli from the psycholinguistic study of~\citet{ferguson2008anomalies}. There are 128 items from the original psycholinguistic experiments, and we synthetically generate 10,720 additional items (see Appendix~\ref{data_generation} for illustration of data generation process). 
We match target nouns and syntactic constructions across conditions in order to control lexical properties that influence language models’ predictions. Table~\ref{tab:stimuli_common} shows example items from the synthetic large-scale dataset (see Section \ref{sec:small-scale} for example items from the small-scale dataset). 

\begin{table}[!htb]
   \centering  
    \resizebox{.45\textwidth}{!}{%
    \begin{tabular}{p{.1\linewidth}p{\linewidth}}\toprule
    \textbf{Cond} & \textbf{Sentence}\\\midrule
    CW & If \red{cats} were \green{vegetarians}, people would love them. Families would feed \red{cats} with \emph{\red{fish}/\underline{\green{cabbages}}}.\\\midrule
    RW & Because \red{cats} are \red{carnivores}, people love them. Families would feed \red{cats} with \emph{\underline{\red{fish}}/\green{cabbages}}.\\\midrule
    BB &  Families would feed \red{cats} with \emph{\underline{\red{fish}}/\green{cabbages}}\\\bottomrule
    \end{tabular}
    }
    \caption{Exp1 items (logical completion underlined).\label{tab:stimuli_common}}
\end{table}
\begin{figure}[!thb]
    \centering
    \includegraphics[width = .9\linewidth]{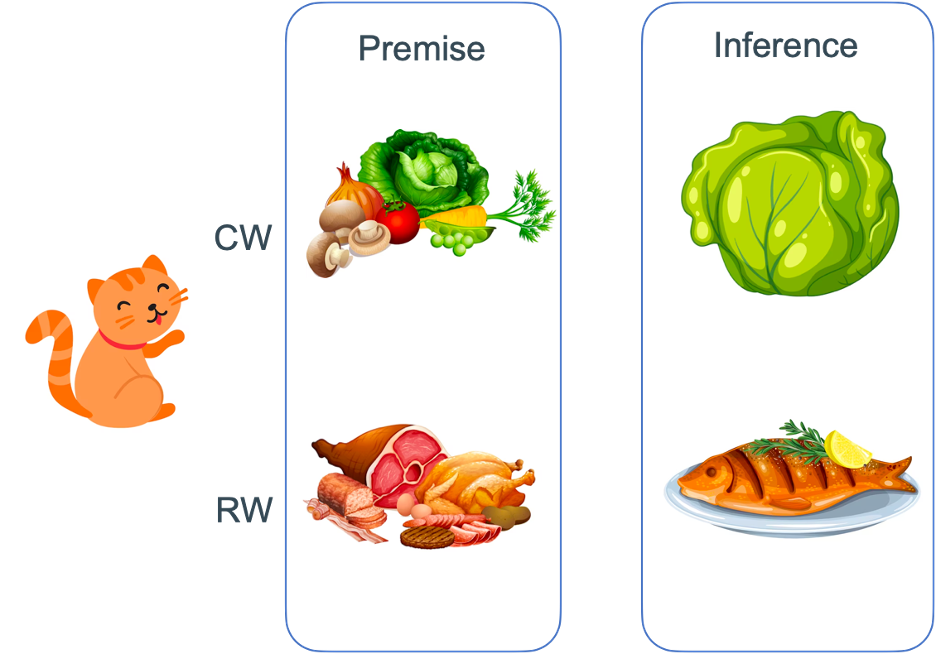}
    \caption{Illustration of Exp1 set-up}
    \label{fig:exp1}
\end{figure}

The experiment includes two key conditions: Counterfactual-World (CW) and Real-World (RW) (Fig.~\ref{fig:exp1}). The CW condition presents a counterfactual scenario, e.g., in which cats are vegetarians. The logical target completion in this example is ``cabbages'', but because in reality cats are more likely to eat fish, this contradicts world knowledge. By contrast, in the RW condition the logical completion is consistent with the real world (``feed cats with fish''). We also include one Baseline Bias (BB) condition, for a more direct test of the strength of models' baseline preference for each completion.

\paragraph{Experiments}
We test counterfactual reasoning in five pre-trained language models. We include autoregressive transformers in the GPT family (GPT-2~\citep{radford2019language} and GPT-3~\citep{brown2020language}) and masked language models in the BERT family (BERT~\citep{devlin-etal-2019-bert}, RoBERTa~\citep{liu2019roberta} and MPNet~\citep{song2020mpnet})~\footnote{We used the smallest uncased variants of GPT-2, BERT, RoBERTa, and MPNet, and we used the text-davinci-003 variant of GPT-3 via API request. Experiments were conducted from April to August 2022.}.

We test models by comparing the log-probability that each model assigns to the CW-congruent (``cabbages'') and RW-congruent (``fish'') completions given the contexts. For all conditions, we compute the percentage of items in which the CW-congruent continuation has a higher probability than the RW-congruent continuation. This means that in RW and BB conditions, \emph{lower} values reflect better predictions, since the CW-congruent completion is the less logical completion in these conditions. 

\paragraph{Results}
\begin{table}[!htb]
    \centering
    \resizebox{.45\textwidth}{!}{%
    \begin{tabular}{ccccc@{\hspace{0.05cm}}ccc}
    \toprule
    \multirow{2}{*}{Model} & \multicolumn{3}{c}{Small-scale}& & \multicolumn{3}{c}{Large-scale} \\ \cmidrule{2-4} \cmidrule{6-8}
     & CW & RW  & BB && CW & RW & BB \\\midrule
    GPT2 &  53.1 & 34.4 & 40.6 && 53.7 & 29.5 & 31.5 \\
    GPT3 & \textbf{68.8} & \textbf{18.8} & \textbf{18.7} && \textbf{71.3} & \textbf{2.5} & \textbf{14.7} \\
    BERT & 46.9 & 43.8 & 31.2 && 34.2 & 14.3 & 35.2 \\
    RoBERTa & 53.1 & 21.9 & 21.9 && 61.4 & 26.5 & 47.2 \\
    MPNet & 50.0 & 21.9 & 21.9 && 66.9 & 15.6 & 36.6\\\bottomrule
    \end{tabular}%
    }
    \caption{Percentage of preference for CW-congruent completion (e.g., ``cabbages'') in Exp1. In the CW condition, \emph{higher} values reflect better predictions. In RW and BB conditions, \emph{lower} values reflect better predictions. \label{tab:result_common}}
\end{table}

Table~\ref{tab:result_common} shows the preferences for CW-congruent completions across all models and conditions, for the small-scale hand-designed items from the psycholinguistic experiment, and for the large-scale synthetic items.~\footnote{A potential concern with aggregated percentages shown in Table~\ref{tab:result_common} and Table~\ref{tab:result_context} is that given a specific instance, a model may assign a higher probability to a CW-congruent continuation in the CW condition because it incorrectly predicts the corresponding BB/RW item. This concern is mitigated by the fact that we focus our conclusions on the difference between the CW and RW conditions, rather than the accuracies in the CW condition alone. However, to further address this concern, we calculate the proportion of items in which the model shows the correct preference in both CW and RW conditions. The results are presented in Section \ref{sec: by-item} and suggest a comparable pattern in terms of relative model strengths.} We see that all models show stronger preference for CW-congruent continuations in the counterfactual (CW) context than in the other conditions (though in the case of BERT on the small-scale data, this difference is negligible). All models show below-chance preference for CW-congruent continuations in the RW condition---which means above-chance preference for the correct RW-congruent continuations. However, though all model preferences for the correct CW-congruent continuation are higher in the CW condition than in the RW condition, even in the CW condition the preference for CW-congruent conditions is at best slightly above chance for most models. The exception is GPT-3, which is the only model to prefer the CW-congruent continuation in greater than 70\% of items.

We also see that GPT-3 shows exceptionally strong performance on both BB and CW conditions. This suggests, slightly counterintuitively, that stronger grasp of relevant world knowledge may in fact be associated with models \emph{more} effectively overriding that knowledge in a counterfactual. To investigate this effect further, we examine the impact of world knowledge at the item level. We quantify strength of world knowledge as the difference between models' log-probability of CW- and RW-congruent continuations for a given item in the BB condition, and the strength of counterfactual preference as the difference between log-probability of CW- and RW-congruent continuations for a given item in the CW condition. We then compute the Pearson correlation between these strength measures. We find a significant correlation between the robustness of world knowledge encoding and strength of counterfactual preference in the CW condition across all language models (see Appendix~\ref{app:correlation}), further supporting a relationship between strength of world knowledge and counterfactual sensitivity. While previous work has suggested that large language models may have difficulty avoiding memorized texts when explicitly prompted to end famous quotes differently~\citep{mckenzie2022inverse}, our results suggest that world knowledge may in fact facilitate reasoning when accompanied with clear structural cues (e.g. ``if''). To better understand how world knowledge informs language models' predictions and inference, it will be important to continue expanding the scale of tests and more carefully operationalize definitions of world knowledge in future work.

\section{Exp2: impact of cue words in context}\label{sec: cue_word}

The first experiment suggests that models can to an extent override world knowledge given a counterfactual, particularly in cases when models have a strong handle on the relevant world knowledge. However, it is possible that in these tests the models were not relying on sophisticated understanding of counterfactuals, but rather on simple lexical triggers in context. Consider, for instance, that models could perform well in Exp1 if they simply increase their preference for ``cabbages'' in the proximity of ``vegetarians'', etc. To test the impact of these lexical triggers, we incorporate an additional condition.

\begin{figure}[!thb]
    \centering
    \includegraphics[width=0.9\linewidth]{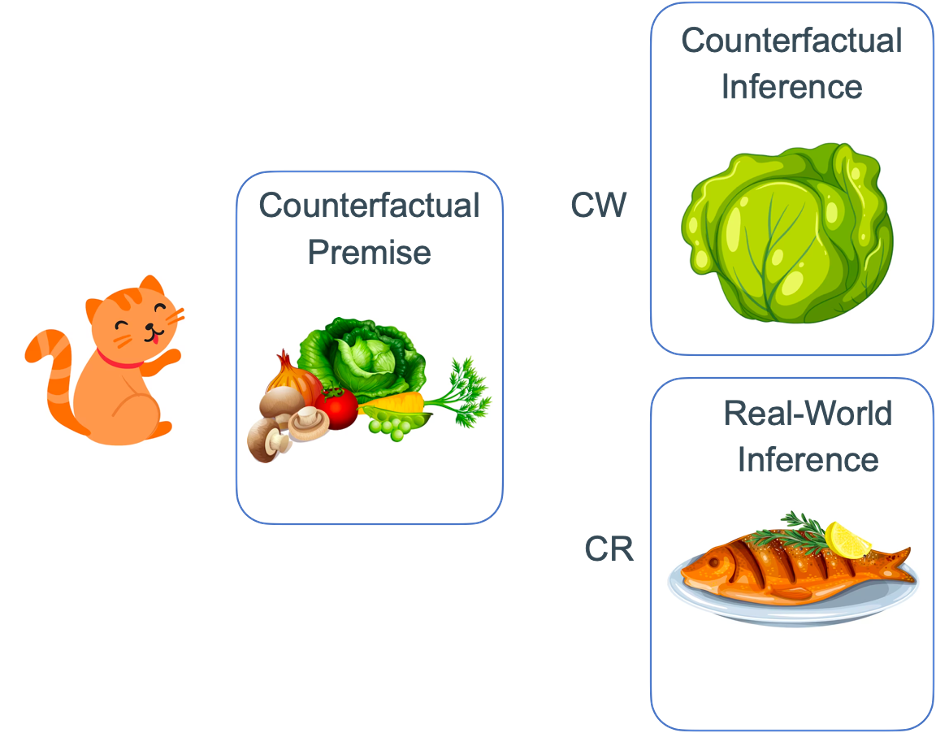}
    \caption{Illustration of Exp2 set-up.}
    \label{fig:exp2}
\end{figure}

\paragraph{Items}
Table~\ref{tab:stimuli_common_baseline} and Fig.~\ref{fig:exp2} show a sample item and illustration of experimental set-up with the new added condition. In this Counterfactual-to-Reality (CR) condition, models see the same counterfactual context, but the subsequent sentence references actual reality. So the correct completion is consistent with reality, but inconsistent with the lexical trigger (``vegetarians''). We generate sentences in the CR condition by modifying CW sentences to include the discourse connective ``In reality'' and to include present tense in the second sentence.

\begin{table}[!htb]
    \centering
    \resizebox{.45\textwidth}{!}{%
    \begin{tabular}{p{.1\linewidth}p{\linewidth}}
    \toprule
    \textbf{Cond} & \textbf{Sentence}\\\midrule
    CR & If \red{cats} were \green{vegetarians}, people would love them. \textbf{In reality}, families feed \red{cats} with \emph{\underline{\red{fish}}/\green{cabbages}}.\\\bottomrule
    \end{tabular}
    }
    \caption{Exp2 items (logical completion underlined).\label{tab:stimuli_common_baseline}}
\end{table}
\paragraph{Experiments}
As above, we calculate percentage of items in which models prefer the CW-congruent continuations. Models relying on information beyond simple lexical triggers should show a sharp drop in preference for the CW-congruent completion in the CR condition, where the correct completion should align with real world information.

\paragraph{Results}
Table~\ref{tab:result_common_sensitivity} shows the results. We see that most models show non-zero drop between CW and CR conditions---however, for most models this reduction is minor. It is only GPT-3 that shows a truly substantial drop in CW-congruent preference, and only in the large-scale synthetic dataset. This suggests that most models are largely following simpler lexical triggers, while GPT-3 has somewhat greater sensitivity to more detailed linguistic cues. Note, however that GPT-3's relative success on the synthetic data over the small-scale data may rely on larger distance between lexical triggers and target positions: see Appendix~\ref{app:ablation} for evidence on GPT-3's sensitivity to linear distance.

\begin{table}[!htb]
\centering
\resizebox{.34\textwidth}{!}{%
\begin{tabular}{cccc@{\hspace{0.05cm}}cc}
\toprule
\multirow{2}{*}{Model} & \multicolumn{2}{c}{Small-scale} & & \multicolumn{2}{c}{Large-scale} \\\cmidrule{2-3} \cmidrule{5-6}
 & CW & CR && CW & CR \\ \midrule
GPT2 & 53.1 & 50.0 && 53.7 & 51.9 \\
GPT3 & \textbf{68.8} & 56.2 && \textbf{71.3} & \textbf{28.0}\\
BERT & 46.9 & 46.9 && 34.2 & 39.4 \\
RoBERTa & 53.1 & \textbf{37.5} && 61.4 & 57.3\\
MPNet & 50.0 & 46.9 && 66.9 & 58.1 \\\bottomrule
\end{tabular}%
}
\caption{Percentage of preference for CW-congruent completion (e.g., ``cabbages'') in Exp2. In the CW condition, \emph{higher} values reflect better predictions. In the CR condition, \emph{lower} values reflect better predictions.\label{tab:result_common_sensitivity}}
\end{table}
\section{Exp3: Inferring real world state with counterfactual cues}\label{sec:exp3}

The previous experiments indicate that models can override world knowledge in the face of counterfactual evidence, and that the ability to do this improves with stronger world knowledge---but for most models this performance appears to be driven largely by simple lexical triggers in the context, with the possible exception of GPT-3. In this section we remove the influence of pre-existing world knowledge, and hold constant lexical triggers across conditions, for a more direct test of models' sensitivity to linguistic indicators of counterfactuals, and what they say about the true state of the world. This task is particularly challenging because language models must infer the true state of the world based on the presence of counterfactuals, with lexical cues often being misleading.

\paragraph{Items}
We adapt stimuli from a psycholinguistic study with 96 controlled sentences \citep{ferguson2012eye}. We additionally create a larger-scale synthetic dataset with 12,960 sentences, using the same events as the generated dataset from Section~\ref{overriding-world-knowledge}. We modify the subject noun phrases such that there is no influence of existing world knowledge. For example, we modify the subject ``cat'' to ``pet'', so that there is no prior knowledge about the subject's preference for ``cabbages'' or ``fish''.
As a result, existing world knowledge cannot inform the correct completion---instead, models need to infer based on the counterfactual language that the true state of the world is different from what the counterfactual states. Further, we control the lexical items used across different conditions to minimize effects of lexical cues on condition differences (see Table~\ref{tab:stimuli_context}). 

\begin{table}[!htb]
    \centering
    \resizebox{.45\textwidth}{!}{%
    \begin{tabular}{p{.11\linewidth}p{\linewidth}}\toprule
    \textbf{Cond} & \textbf{Sentence}\\\midrule
    CWC & If \brown{the pets} were \green{vegetarians}, people would love them. In fact, people feed \brown{the pets} with \emph{\underline{\red{fish}}/\green{cabbages}}.\\\midrule
    RWCA & Because \brown{the pets} are \green{vegetarians}, people love them. In fact, people feed \brown{the pets} with \emph{\red{fish}/\underline{\green{cabbages}}}.\\\midrule
    BBC &  In fact, people feed \brown{the pets} with \emph{\red{fish}/\green{cabbages}}.\\\bottomrule
    \end{tabular}
    }
    \caption{Exp3 items (logical completion underlined).\label{tab:stimuli_context}}
\end{table}
\begin{figure}[!thb]
    \centering
    \includegraphics[width = 0.9\linewidth]{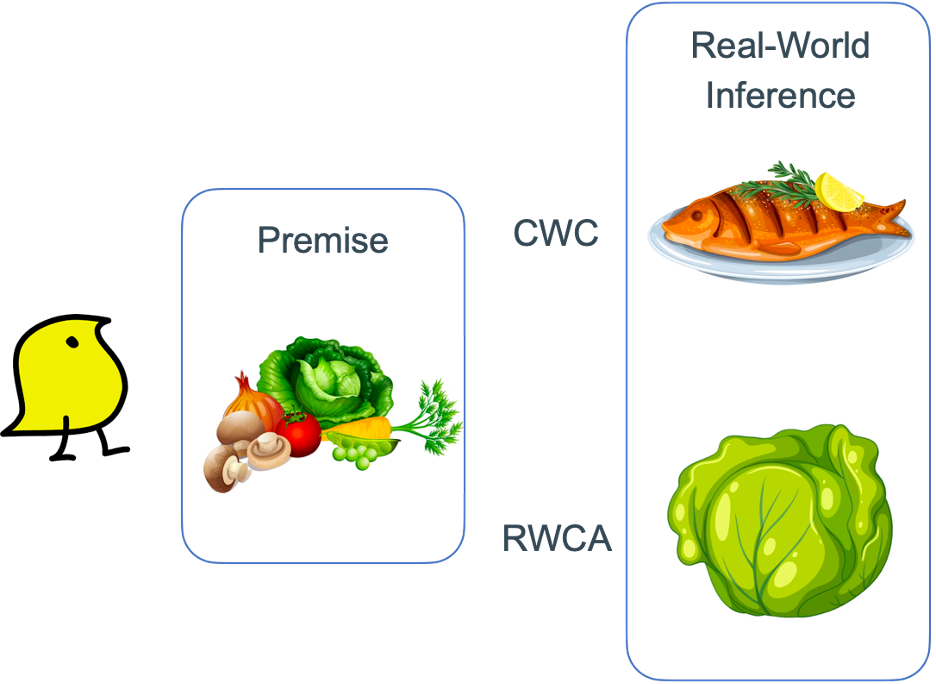}
    \caption{Illustration of Exp3 set-up.}
    \label{fig:exp3}
\end{figure}
Fig.~\ref{fig:exp3} shows the set-up of conditions. In the Counterfactual-World Context (CWC) condition, the scenario described in the first sentence is neutral with respect to real world knowledge---it is the use of the counterfactual (``if...were'') that tips us off that this scenario is not true in reality. The correct completion, then, cannot be informed by world knowledge, and is also misaligned with the lexical trigger (e.g., ``vegetarians''), so models must rely specifically on this implication from the counterfactual in order to perform well. 

In the Real-World Context Alternative (RWCA) condition, the context uses the same lexical triggers (``vegetarians'') as the CWC condition. However, because there is no counterfactual language, the logical completion is now the word associated with the lexical trigger (e.g., ``cabbages'', associated with ``vegetarians''). 

Given that the logical completions in CWC and RWCA differ, we also compare against a Baseline Bias Context (BBC) condition, to establish default model preference for the target factual completion in the presence of the new subject noun phrase. 

\paragraph{Experiments}
We compare proportion of CWC-congruent completions across conditions. Good performance will assign high values in the CWC condition and low values in the RWCA condition.

\paragraph{Results}
\begin{table}[!htb]
\centering
\resizebox{.45\textwidth}{!}{%
\begin{tabular}{ccccc@{\hspace{0.05cm}}ccc}
\toprule
\multirow{2}{*}{Model} & \multicolumn{3}{c}{Small-scale}& & \multicolumn{3}{c}{Large-scale} \\ \cmidrule{2-4} \cmidrule{6-8}
 & CWC & RWCA  & BBC && CWC & RWCA & BBC \\\midrule
GPT2 & 66.7 & 66.7 & 33.3 && 35.8 & 32.2 & 72.6 \\
GPT3 & \textbf{62.5} & \textbf{33.3} & 50.0 && \textbf{47.6} & \textbf{32.2} & 73.8 \\
BERT & 45.8 & 33.3 & 50.0 && 53.0 & 53.0 & 71.5 \\
RoBERTa & 50.0 & 50.0 & 50.0 && 35.7 & 31.3 & 72.5 \\
MPNet & 37.5 & 33.3 & 62.5 && 41.4 & 32.3 & 68.5\\\bottomrule
\end{tabular}%
}
\caption{Percentage of preference for CWC-congruent completion (e.g., ``fish'') in Exp3. In the CWC condition, \emph{higher} values reflect better predictions. In the CWCA condition, \emph{lower} values reflect better predictions. The BBC condition establishes models' default preference for the CWC-congruent completion. \label{tab:result_context}}
\end{table}
Table~\ref{tab:result_context} shows the results. In the small-scale dataset, most models show a similar preference in CWC and RWCA, suggesting again that their predictions are largely driven by lexical triggers. Only GPT-3 shows substantial difference between CWC and RWCA, indicating finer-grained sensitivity to counterfactual structures. This sensitivity is, however, less pronounced in the large-scale dataset. Closer inspection suggests that GPT-3's specific success on the small-scale data may in fact be attributable to canceling out of lexical triggers: in the small-scale dataset, there are lexical triggers supporting both continuations (see \ref{sec:small-scale} for more illustration of the characteristics of the small-scale dataset), which may cause lexical cues to cancel out, enabling more influence from other linguistic cues. To take one example, the small-scale dataset contains the item ``If Helen had received her \underline{student loan}, her bank balance would now be \underline{in credit}. When she checked her bank balance she was \textbf{worried/happy} about her finance.'' In this item, among the lexical triggers (``student loan'', ``in credit'', ``bank balance'') there are potential associations with both the CWC-congruent completion ``worried'' and the CWC-incongruent completion ``happy''. By contrast, in the large-scale dataset, the major lexical trigger (``vegetarians'') always favors the CWC-incongruent continuation (``cabbages''), causing strong lexical bias against the CWC-congruent continuation (see Appendix~\ref{app:ablation} for further analysis on the role of conflicting lexical triggers and other linguistic factors). This suggests that GPT-3 does show real sensitivity to linguistic indicators of counterfactuals, but the effect of superficial lexical cues remains strong.

\section{Conclusion}
The experiments above have shown that when presented with counterfactual situations, PLMs are able to prefer completions that conflict with world knowledge---and counterintuitively, this sensitivity appears better in cases where that world knowledge is stronger. Our results also indicate, however, that models are in large part relying on simple lexical cues to inform these preferences. The only model that shows more sophisticated sensitivity to fine-grained linguistic cues separating counterfactuals from reality is GPT-3---which successfully distinguishes conditions based on counterfactual cues, but nonetheless still shows strong influences from lexical associative cues. Why might world knowledge aid counterfactual sensitivity? Does GPT-3 truly understand counterfactuals? One possibility worth considering is that explanations in both of these cases involve volume of exposure. First, models' stronger world knowledge for a given fact suggests that models have encountered that fact more often in training---and this may in turn translate to more exposure to that type of knowledge in counterfactual contexts, enabling more straightforward memorization-based performance. Similarly, while GPT-3 may robustly understand counterfactuals, the massive data exposure for that model may enable a simpler path to success: GPT-3 could simply have developed lower-level knowledge of how linguistic cues like ``If/had'' versus ``Because'' mediate levels of association between nearby lexical cues and later words. We leave investigation of these hypotheses for future work.

\section*{Limitations}
The datasets in this paper systematically control lexical cues and world knowledge between critical conditions, allowing us to tease apart the effects of statistical heuristics versus reasoning about causal relations. However, the manipulation brings unnaturalness to sentences when scaling up into large-scale synthetic datasets, and constrains the level of linguistic complexity. As we have seen in Exp3, the small-scale dataset has more complex combinations of conflicting lexical triggers than the large-scale dataset, causing language models to behave differently across datasets. Though we further address the effects of conflicting lexical cues in Appendix~\ref{app:ablation}, it will be valuable to carry out additional investigation of effects of sentence naturalness, and to consider designing large-scale datasets using naturally-occurring data.

The study raises and leaves open a number of interesting questions: How exactly might counterfactual reasoning benefit from world knowledge? To what extent does GPT-3's stronger performance reflect robust logical and counterfactual reasoning? While we lay out some possible explanations in the Conclusion and investigate the role of other linguistic and non-linguistic factors in the above experiments and in the Appendix, we leave additional systematic analysis for future work.

Finally, the experiments use English, in which counterfactual conditionals have distinct and systematic linguistic markers relative to other types of conditionals. It would be interesting to investigate other languages in which counterfactual conditionals are not marked linguistically, and require world knowledge to disambiguate. For example, a Chinese conditional could be ambiguous between ``if it had rained today'' and ``if it rains today''.

\section*{Ethics Statement}
The datasets were either created and published by reseachers in psycholinguistics, or synthetically generated by the authors without use of harmful information. No experiments involving human subjects were included in the paper. The authors do not foresee any ethical concerns in this paper.

\bibliography{anthology,custom}
\bibliographystyle{acl_natbib}

\appendix

\section{Appendix}

\subsection{Example items in small-scale dataset}\label{sec:small-scale}
Table \ref{tab:exp1_small} shows the example items from Exp1 and Exp2 in the small-scale psycholinguistic datasets, and Table \ref{tab:exp3_small} shows the example items from Exp3 in the small-scale dataset. Semantic association between the target word and key lexical items in the context is less salient (e.g. ``language skills'' and ``talk'') in the small-scale dataset as compared to the association in the large synthetic dataset (e.g. ``vegetarian'' and ``carrots''). In particular, sentences in Exp3 contain lexical triggers that could support both CWC-congruent and RWCA-congruent continuations. For instance, the key lexical items (``student loan'', ``bank balance'', ``in credit'') could be logically associated with either of the feelings (``happy'' or ``worried''). 

\begin{table}[!htb]
    \centering
    \resizebox{.45\textwidth}{!}{%
    \begin{tabular}{p{.11\linewidth}p{\linewidth}}\toprule
    \textbf{Cond} & \textbf{Sentence}\\\midrule
    CW & If \red{cats} had developed \green{language skills} like humans it would be interesting to hear what they have to say. Judith would listen to her cat \textit{\red{meow}/\underline{\green{talk}}} and throw balls of wool for it to play with. \\\midrule
    RW & If \red{cats} are bored and want something to do they are usually very good at letting their owners know. Judith would listen to her cat \textit{\underline{\red{meow}}/\green{talk}} and throw balls of wool for it to play with. \\\midrule
    BB &  Judith would listen to her cat \textit{\underline{\red{meow}}/\green{talk}} and throw balls of wool for it to play with.\\
    CR & If \red{cats} had developed \green{language skills} like humans it would be interesting to hear what they have to say. \textbf{In reality}, Judith listens to her cat \textit{\red{meow}/\underline{\green{talk}}} and throws balls of wool for it to play with. \\\bottomrule
    \end{tabular}
    }
    \caption{Example Exp1 and Exp2 items in small-scale dataset (logical completion underlined).\label{tab:exp1_small}}
\end{table}

\begin{table}[!htb]
    \centering
    \resizebox{.45\textwidth}{!}{%
    \begin{tabular}{p{.11\linewidth}p{\linewidth}}\toprule
    \textbf{Cond} & \textbf{Sentence}\\\midrule
    CWC & If Helen had received her first student loan, her bank balance would now be in credit. When she checked her bank balance today she was \textit{\underline{\red{worried}}/\green{happy}} with her financial situation. \\\midrule
    RWCA & Because Helen had received her first student loan, her bank balance was now in credit. When she checked her bank balance today she was \textit{\red{worried}/\underline{\green{happy}}} with her financial situation. \\\midrule
    BBC &  When she checked her bank balance today she was \textit{\red{worried}/\green{happy}} with her financial situation. \\\bottomrule
    \end{tabular}
    }
    \caption{Example Exp3 items in small-scale dataset (logical completion underlined).\label{tab:exp3_small}}
\end{table}

\label{sec:appendix}

\subsection{Generation process of dataset}\label{data_generation}
We design our synthetic dataset to parallel the psycholinguistic stimuli. We design a series of causal event pairs (e.g. ``like''/``feed''), and situate these pairs within counterfactual conditional templates (e.g. “if {subject1} liked {object1}, {subject2} would feed {subject1} with {object2}”). For each subject/object slot, we define a class of nouns satisfying both selection restriction of the verb and world knowledge. For example, in the template sentence "if {subject1} liked vegetables, families would feed them with \emph{cabbages/chicken}", subject1 can be carnivores (e.g. ``cats/lions/tigers''). We then vary lexical items in each subject/object slot, and other linguistic markers (e.g. modal, tense) in the template. Table~\ref{tab:data_generation} shows examples illustrating the data generation from a sample event in the CW-condition in Exp1. Exp2 and Exp3 use the same template and we manipulate the syntactic structure or informativity of the subject as described in Section~\ref{sec: cue_word} and Section~\ref{sec:exp3}.

\begin{table}[!htb]
    \centering
    \resizebox{.45\textwidth}{!}{%
    \begin{tabular}{p{.2\linewidth}p{0.75\linewidth}}\toprule
    \textbf{Condition} & \textbf{Sentence}\\\midrule
    Original & If cats were vegetarians, families would feed them with cabbages.\\
    $\mathrm{Subject_{1}}$ & If \emph{dogs} were vegetarians, families would feed them with cabbages.\\
    $\mathrm{Object_{1}}$ &  If cats were \emph{greens}, families would feed them with cabbages.\\
    $\mathrm{Subject_{2}}$ & If cats were vegetarians, \emph{breeders} would feed them with cabbages.\\
    $\mathrm{Object_{2}}$ & If cats were vegetarians, breeders would feed them with \emph{cabbages}.\\
    Modal & If cats were vegetarians, families \emph{might} feed them with cabbages.\\
    Tense & If cats \emph{had been} vegetarians, families would \emph{have fed} them with cabbages.\\\bottomrule
    \end{tabular}
    }
    \caption{Illustration of data generation process in large-scale synthetic dataset. Different sentences can be generated based on the \emph{original} sentence by changing lexical items in subject and object positions.}
    \label{tab:data_generation}
\end{table}

\subsection{Correlation with world knowledge}\label{app:correlation}
Table~\ref{tab:result_correlation} shows the correlation between the robustness of world knowledge representation and the strength of counterfactual preference in CW condition. Across all language models there is a significant correlation, with all correlation coefficients at or above 0.69, indicating that language models benefit from a good representation of world knowledge for this counterfactual task. 
\begin{table}[!htb]
    \centering
    \resizebox{.45\textwidth}{!}{%
    \begin{tabular}{ccccc@{\hspace{0.05cm}}ccc}
    \toprule
    \multirow{2}{*}{Model} & \multicolumn{2}{c}{Small-scale}& & \multicolumn{2}{c}{Large-scale} \\ \cmidrule{2-3} \cmidrule{5-6}
     & \emph{coef} & \emph{p} && \emph{coef} & \emph{p}  \\\midrule
    GPT2 &  .86 & $<$.001***  && .82 & $<$.001***  \\
    GPT3 & .70 & .004**  && .74 & $<$.001***  \\
    BERT & .91 & .001** && .69 & $<$.001***  \\
    RoBERTa & .86 & .001***  && .77 & $<$.001***  \\
    MPNet & .88 & $<$.001*** && .61 & $<$.001*** \\\bottomrule
    \end{tabular}%
    }
    \caption{Correlation between robustness of world knowledge encoding and strength of counterfactual preference in CW condition. $p < .05$*, $p < .01$**, $p < .001$***. \label{tab:result_correlation}}   
\end{table}

\subsection{Follow-up analysis on GPT-3's success}\label{app:ablation}
The previous experiments indicate that GPT-3 has the best performance in counterfactual tasks. We also find that GPT-3's success differs non-trivially between small-scale and large-scale datasets. In Exp2, GPT-3 is more successful on the large-scale dataset. By contrast, in Exp3, GPT-3 is more successful on the small-scale dataset. What kind of linguistic factors are driving the success of GPT-3? Why is there an asymmetry between GPT-3's performance on the small-scale and large-scale datasets? We speculate that there are two possible reasons related to the design and characteristics of the small-scale and large-scale datasets. First, the linear distance between lexical triggers and target positions in the large-scale dataset is not controlled as carefully as in the small-scale dataset. Second, lexical triggers in the large-scale dataset always favor a specific continuation, whereas in the small-scale dataset the cue can support both continuations.

In this experiment, we further explore these questions by investigating to what extent GPT-3's success relies on other linguistic factors. We first design a \emph{Baseline} dataset by selecting a subset of the large-scale dataset from Exp3, with the criterion that the selected items have no strong bias toward either completion in the Baseline Bias Context (BBC) condition (see examples in Table~\ref{tab:cue_test}). Next, we test GPT-3's sensitivity to three classes of linguistics features: conflicting lexical triggers, linear distance to target position, and several other linguistic markers. We manipulate these linguistic features in the items of the CWC and RWCA conditions, to form three new datasets. The \emph{Cue} dataset introduces a conflicting cue via a discourse connective ``rather than'' (see examples in Table~\ref{tab:cue_test_rather}). The \emph{Distance} dataset changes the position of the conflicting lexical cues by using the discourse connective ``instead of'' (see examples in Table~\ref{tab:cue_test_instead}). The \emph{Marker} dataset manipulates other fine-grained linguistic markers including sentence boundary, tense, discourse connective (see examples in Table~\ref{tab:cue_test_markers}). There are 10,000 items in total. We again calculate percentage of items in which the model prefers CWC-congruent continuations.

 
\paragraph{Baseline} We test GPT-3's preference for CWC-congruent continuations in the \emph{Baseline} dataset to establish a baseline comparison for subsequent analysis. The results are shown in the right-hand column of Table~\ref{tab:cue_test}. Similar to the results in Section~\ref{sec:exp3}, GPT-3 shows a greater preference for CWC-congruent continuations in the CWC condition than in the RWCA condition, even when there is not a strong preference in the BBC condition, which indicates GPT-3's sensitivity to counterfactual structure. 

\begin{table}[!htb]
    \centering
    \resizebox{.45\textwidth}{!}{%
    \begin{tabular}{p{.18\linewidth}p{0.7\linewidth}p{0.14\linewidth}}
     \toprule Condition & Sentence & GPT-3 \\\midrule
      CWC  &  If \brown{the pet} had loved \green{vegetables}, it would be very surprising. In fact, people feed \brown{the pet} with \emph{\underline{\red{fish}}/\green{cabbages}}. & 34.8\\
      RWCA & Because \brown{the pet} loved \green{vegetables}, it was very surprising. In fact, people feed \brown{the pet} with \emph{\red{fish}/\underline{\green{cabbages}}}. & 27.3\\
      BBC &  In fact, people feed \brown{the pet} with \emph{\red{fish}/\green{cabbages}}. & 42.5\\\bottomrule
    \end{tabular}
    }
        \caption{\emph{Baseline} dataset: Example items and percentage of preference for CWC-congruent completion (e.g., ``fish'').\label{tab:cue_test}}
\end{table}

\paragraph{Conflicting lexical cue} Next, in the \emph{Cue} dataset we test to what extent GPT-3's performance reflects canceling out of lexical cues, by adding a conflicting lexical cue to the context, licensed by the discourse connective ``rather than''. Though a new conflicting lexical cue appears, the logical completion should remain the same. Table~\ref{tab:cue_test_rather} (right-hand column) shows that GPT-3 is greatly affected by the presence of conflicting lexical cues. After inserting the conflicting cue (e.g., ``meat'') into context, the percentage of CWC-congruent continuations (e.g., ``fish'') increased in both CWC and RWCA conditions, indicating a strong effect from the presence of a conflicting lexical cue.

\begin{table}[!thb]
    \centering    
    \resizebox{.45\textwidth}{!}{
    \begin{tabular}{p{.18\linewidth}p{0.7\linewidth}p{0.14\linewidth}}
     \toprule Condition & Sentence & GPT-3 \\\midrule
      CWC (Rather)  &  If \brown{the pet} had loved \green{vegetables} rather than \red{meat}, it would be very surprising. In fact, people feed \brown{the pet} with \emph{\underline{\red{fish}}/\green{cabbages}}. & 48.5\\
      RWCA (Rather) & Because \brown{the pet} loved \green{vegetables} rather than \red{meat}, it was very surprising. In fact, people feed \brown{the pet} with \emph{\red{fish}/\underline{\green{cabbages}}}. & 47.0\\\bottomrule
    \end{tabular}
    }
    \caption{\emph{Cue} dataset: Example items and percentage of preference for CWC-congruent completion (e.g., ``fish''). \label{tab:cue_test_rather}}
\end{table}

\paragraph{Linear distance to target}
Next, we use the \emph{Distance} dataset to test the extent to which the salience of lexical cues is affected by distance from the target word. To do this, we move the conflicting lexical cues to the beginning of the sentence, using the discourse connective ``instead of''. As a result, the conflicting cue (e.g. ``meat'') is moved farther away from the target, compared to it in \emph{Cue} dataset. Table~\ref{tab:cue_test_instead} (right-hand column) shows the results. The model is less likely to predict the CWC-congruent continuation (e.g., ``fish'') in both conditions. The result suggests that linear distance from lexical cues to the prediction target has a strong impact.

\begin{table}[!htb]
    \centering
    \resizebox{.45\textwidth}{!}{
    \begin{tabular}{p{.18\linewidth}p{0.7\linewidth}p{0.14\linewidth}}
     \toprule Condition & Sentence & GPT-3 \\\midrule
      CWC (Instead)  &  If instead of \red{meat}, \brown{the pet} had loved \green{vegetables}, it would be very surprising. In fact, people feed \brown{the pet} with \emph{\underline{\red{fish}}/\green{cabbages}}. & 28.5\\
      RWCA (Instead) & Because instead of \red{meat}, \brown{the pet} loved \green{vegetables}, it was very surprising. In fact, people feed \brown{the pet} with \emph{\red{fish}/\underline{\green{cabbages}}}. & 33.8\\
    \end{tabular}
    }
    \caption{\emph{Distance} dataset: Example items and percentage of preference for CWC-congruent completion (e.g., ``fish'').\label{tab:cue_test_instead}}
\end{table}

\paragraph{Other linguistic markers} Finally, we use the \emph{Marker} dataset to probe how other fine-grained linguistic markers affect the accuracy of predictions in counterfactual sentences. We test the effect of sentence boundaries (indicated by a period), discourse connectives (indicated by ``In fact'') and tense. All three manipulations make CWC-congruent continuations less coherent relative to the CWC condition in the \emph{Baseline} dataset, while the tense and sentence boundary manipulations additionally cause the RWCA-congruent continuation to become more logical. Table~\ref{tab:cue_test_markers} (right-hand column) shows the results. GPT-3 shows a fair amount of sensitivity to these linguistic markers. For the linguistic markers (tense marker, sentence boundary marker) that shift the logical completion from CWC-congruent (e.g. ``fish'') to RWCA-congruent (e.g. ``cabbages''), GPT-3 is less likely to prefer the CWC-congruent completion, with tense generating the strongest effect. For the discourse connective manipulation, which deletes the connective ``in fact'', and should decrease the preference for the CWC-congruent completion, GPT-3 instead shows a slightly stronger preference for those CWC-congruent completions. 

\begin{table}[!thb]
    \centering
   \resizebox{.45\textwidth}{!}{
    \begin{tabular}{p{.18\linewidth}p{0.7\linewidth}p{0.14\linewidth}}     \toprule Condition & Sentence & GPT-3 \\\midrule
      Boundary  &  If \brown{the pet} had loved \green{vegetables}, it would be very surprising\textbf{\textcolor{blue}{,}} in fact, people feed \brown{the pet} with \emph{\red{fish}/\underline{\green{cabbages}}}. & 28.7\\
      Connective & If \brown{the pet} loved \green{vegetables}, it would be very surprising. People feed \brown{the pet} with \emph{\underline{\red{fish}}/\green{cabbages}}. & 35.5\\
      Tense & If \brown{the pet} had loved \green{vegetables}, it would be very surprising. In fact, people \textcolor{blue}{would} feed \brown{the pet} with \emph{\red{fish}/\underline{\green{cabbages}}}. & 14.0
    \end{tabular}
    }
    \caption{\emph{Marker} dataset: Example items and percentage of preference for CWC-congruent completion (e.g., ``fish'').\label{tab:cue_test_markers}}
\end{table}

\subsection{Additional metrics on small-scale dataset}\label{sec: by-item}
To further evaluate whether models' success on counterfactual inference disentangle with the preference towards a specific continuation, we also conduct by-item analysis on the small-scale datasets, and calculate the proportion of trials in which the model demonstrates a preference for the logical completion in both CW and RW conditions for Exp1, and in both CWC and RWCA conditions for Exp3. Table \ref{tab:additional_analysis} shows the percentage of preference for logical completions in both counterfactual and factual conditions in Exp1 and Exp3. The results are consistent with the findings we report in Section 2 and Section 4. In Exp1, GPT-3, RoBERTa and MPNet show above-chance preference (25\%) for logical continuations in both conditions. In Exp3, only GPT-3 shows substantial preference for logical continuations.

\begin{table}[!thb]
    \centering
\resizebox{.45\textwidth}{!}{%
    \begin{tabular}{cccccc}
    \toprule
      Model   &  GPT2 & GPT3 & BERT & RoBERTa & MPNet\\
      Exp1 (CW + RW)  & 18.8 & \textbf{50.0} & 9.4 & 31.3 & 28.1 \\
      Exp3 (CWC + RWCA)& 0 & \textbf{29.2} & 12.5 & 4.1 & 4.2\\\bottomrule
    \end{tabular}
}
    \caption{Percentage of items in which both counterfactual (CW/CWC) and real scenarios (RW/RWCA) are predicted correctly in Exp1 and Exp3.}
    \label{tab:additional_analysis}
\end{table}

\end{document}